\documentclass[runningheads]{llncs}
\usepackage{graphicx}
\usepackage{float}
\usepackage{amssymb}
\usepackage{amsmath}
\usepackage{mathbbol}
\usepackage{booktabs}
\usepackage{multirow}
\usepackage{siunitx}
\usepackage[usenames, dvipsnames]{color}

\usepackage{changepage}
 \usepackage{algorithm}
 \usepackage{algpseudocode}
 \usepackage{pgf, tikz}
\usetikzlibrary{arrows, automata}
\tikzset{edge/.style = {->,> = latex'}}
\newlength\lengtha 
\newlength\lengthb 

\usetikzlibrary{positioning}

\tikzset{basic/.style={draw,fill=blue!20,text width=1em,text badly centered}}
\tikzset{input/.style={basic,circle}}
\tikzset{weights/.style={basic,rectangle}}
\tikzset{functions/.style={basic,circle,fill=blue!10}}
\begin{document}
\title{Modeling Brain Networks with Artificial Neural Networks}

%
\titlerunning{Artificial Brain Networks}
%
\author{Baran Baris Kivilcim\inst{1} \and
Itir Onal Ertugrul\inst{2} \and
Fatos T. Yarman Vural\inst{1}}
\authorrunning{B. B. Kivilcim et al.}

\institute{
Department of Computer Engineering, Middle East Technical University,  Ankara, Turkey\\
\email{\{baran.kivilcim,vural\}@ceng.metu.edu.tr} \and
Robotics Institute, Carnegie Mellon University, Pittsburgh, PA, USA \\
\email{iertugru@andrew.cmu.edu}
} 

\maketitle              
\begin{abstract}

In this study, we propose a neural network approach to capture the functional connectivities among anatomic brain regions. The suggested approach estimates a set of brain networks, each of which represents the connectivity patterns of a cognitive process. We employ two different architectures of neural networks to extract directed and undirected brain networks from functional Magnetic Resonance Imaging (fMRI) data. Then, we use the edge weights of the estimated brain networks to train a classifier, namely, Support Vector Machines(SVM) to label the underlying cognitive process. We compare our brain network models with popular models, which generate similar functional brain networks. We observe that both undirected and directed brain networks surpass the performances of the network models used in the fMRI literature. We also observe that directed brain networks offer more discriminative features compared to the undirected ones for  recognizing the cognitive processes. The representation power of the suggested brain networks are tested in a task-fMRI dataset of Human Connectome Project and a Complex Problem Solving dataset.


\end{abstract}

\section{Introduction}

Brain imaging techniques, such as, functional Magnetic Resonance Imaging (fMRI) have facilitated the researches to understand the functions of human brain using machine learning algorithms \cite{pereira2009machine,mitchell2004learning,wang2004training,michel2012supervised}.
In traditional approaches, such as Multi-Voxel Pattern Analysis (MVPA), the aim was to discriminate cognitive tasks from the fMRI data itself without forming brain graphs and considering relationship between nodes of graphs. Moreover, Independent Component Analysis (ICA) and Principal Component Analysis (PCA) have been applied to obtain better representations. In addition to feature extraction methods, General Linear Model (GLM) and Analysis of Variance (ANOVA) have been used to select important voxels \cite{pereira2009machine}. None of these approaches take into account the massively connected network structure of the brain \cite{zhou2009detecting,smith2014group,calhoun2009review,calhoun2003ica,mckeown1998independent}. 
Recently, use of deep learning algorithms have also emerged in several studies \cite{kawahara2017brainnetcnn,koyamada2015deep,firat2014deep} to classify cognitive states. Most of these studies mainly focus on using deep learning methods to extract better representations from fMRI data for brain decoding.

Several studies form brain graphs using voxels or anatomical regions as nodes and estimate the edge weights of brain graphs with different approaches. Among them, Richiardi et al. \cite{richiardi2011decoding} have created undirected functional connectivity graphs in different frequency subbands. They have employed Pearson correlation coefficient between responses obtained from all region pairs as edge weights and use these edge weights to perform classification in an audio-visual experiment. Brain graphs, constructed using pairwise correlations and mutual information as edge weights, have been used to investigate the differences in networks of healthy controls and patients with Schizophrenia \cite{lynall2010functional} or Alzheimer's disease \cite{menon2011large,kurmukov2017classifying}. Yet, these studies consider only pairwise relationships while estimating the edge weights and ignore the locality property of the brain.

Contrary to pairwise relationships, a number of studies have estimated the relationships among nodes within a local neighborhood. Ozay et. al. \cite{ozay2012mesh} and Firat et al. \cite{firat2013functional} have formed local meshes around nodes and constructed directed graphs as ensembles of local meshes. They have applied Levinson-Durbin recursion \cite{vaidyanathan2007theory} to estimate the edge weights representing the linear relationship among voxels and have used these weights to classify the category of words in a working memory experiment. Similarly, Alchihabi et al. \cite{alchihabi2018dynamic} have applied Levison-Durbin recursion to estimate the edge weights of local meshes of dynamic brain network for every brain volume in Complex Problem Solving task and have explored activation differences between sub-phases of problem solving. While these studies conserve the locality in the brain, construction of a graph for every time instant discards temporal relationship among nodes of the graph. Onal et al. \cite{onal2015modeling,onal2017new} have formed directed brain graphs as ensemble of local meshes. They have estimated the relationships among nodes within a time period considering the temporal information using ridge regression. Since the spatially neighboring voxels are usually correlated, linear independence assumption of features required for closed form solution to the estimation of linear relationship among voxels is violated. This  may result in large errors and inadequate representation. Since the aforementioned studies form local meshes around each node separately, associativity is ignored in the resulting brain graphs. 

In this study, we propose two brain network models, namely, directed and undirected Artificial Brain Networks to model the relationships among anatomical regions within a time interval using fMRI signals. In both network models, we train an artificial neural network to estimate the time series recorded at node which represent an anatomic region by using the rest of the time series recorded in the remaining nodes. In our first neural network architecture, called directed Artificial Brain Networks (dABN), global relationships among nodes are estimated without any constraint whereas in our second architecture of undirected Artificial Brain Networks (uABN), we apply a weight sharing mechanism to ensure undirected functional connections.

We test the validity of our dABN and uABN in two fMRI datasets and compare the classification performances to the other network models available in the literature. First, we employ the Human Connectome Project (HCP), task-fMRI (tfMRI) dataset, in which the participants were required to complete 7 different mental tasks. The second fMRI dataset contains fMRI scans of subjects solving Tower of London puzzle and has been used to study regional activations of Complex Problem Solving \cite{newman2009fmri,alchihabi2018dynamic}. The task recognition performances of the suggested Artificial Brain Networks are significantly greater than the ones obtained with state of the art functional connectivity methods.


\section{Extraction of Artificial Brain Networks}

In this section, we explain how we estimate the edge weights of directed and undirected brain networks using artificial neural networks. Throughout this study, we represent a brain network by  $G=(V,E)$, where $V = \{ v_{1}, v_{2}, v_{3}, \ldots, v_{M} \}$, denotes the vertices of the network, which represent $M = 90$ anatomical brain regions, $R = \{r_{1}, r_{2}, r_{3}, \ldots, r_{M}\}$. The attribute of each node is the average time series of BOLD activations. The average  BOLD activation of an anatomical region $r_i$ at time $t$  is denoted with $b_{i,t}$. We use all anatomical regions defined by Anatomical Atlas Labeling (AAL) \cite{tzourio2002automated}, except for the ones residing in Cerebellum and Vermis. We represent the edges of the brain network by $ E = \{ e_{i,j} | \forall v_i,v_j \in V, i \neq j \} $. The weights of edges depend on the estimation method. We denote the adjacency matrix which consists of the edge weights, as $ \boldmath{A}$, where $a_{i,j}$ represents the weight of edge from $v_i$ to $v_j$, when the network is directed. When the network is undirected the weight of the edge formed between $v_i$ and $v_j$ is $a_{i,j}= a_{j,i}$. Sample representations of directed and undirected brain networks are shown in Fig. \ref{fig:dir_viz} and Fig. \ref{fig:undir_viz}, respectively.

\begin{figure}
\centering
\begin{minipage}{.5\textwidth}
  \centering
   \begin{tikzpicture}[> = stealth, shorten > = 1pt, auto, node distance = 2cm, semithick]
        \tikzstyle{every state}=[draw = black,thick,fill = white,minimum size = 4mm]
        \node[state] (v0) {$s_0$};
        \node[state] (v1) [below of=v0] {$s_1$};
        \node[state] (v2) [right of=v1] {$s_2$};
        \node[state] (v3) [right of=v0] {$s_3$};
        \draw (v0.north) node[above] {Amygdala};
        \draw (v1.south) node[below] {Cingulum};
        \draw (v2.south) node[below] {Occipital};
        \draw (v3.north) node[above] {Parietal};
        
        \foreach \from/\to in {v0/v1,v0/v2,v0/v3,v1/v2,v1/v3,v2/v3}
    		\draw[edge] (\from) to[bend left=10] (\to);
        \foreach \from/\to in {v0/v1,v0/v2,v0/v3,v1/v2,v1/v3,v2/v3}
            \draw[edge] (\to) to[bend left=10] (\from);
           
 \end{tikzpicture}
  \caption{A Directed Brain Network.}
  \label{fig:dir_viz}
\end{minipage}%
\begin{minipage}{.5\textwidth}
  \centering
 
  \begin{tikzpicture}[
            > = stealth, 
            shorten > = 1pt, 
            auto,
            node distance = 2cm, 
            semithick 
        ]
        \tikzstyle{every state}=[draw = black,thick,fill = white,minimum size = 4mm]
        \node[state] (s0) {$s_0$};
        \node[state] (s1) [below of=s0] {$s_1$};
        \node[state] (s2) [right of=s1] {$s_2$};
        \node[state] (s3) [right of=s0] {$s_3$};
			
        \draw (v0.north) node[above] {Amygdala};
        \draw (v1.south) node[below] {Cingulum};
        \draw (v2.south) node[below] {Occipital};
        \draw (v3.north) node[above] {Parietal};

    \foreach \from/\to in {s0/s1,s0/s2,s0/s3,s1/s2,s1/s3,s2/s3}
    \draw (\from) -- (\to);
 \end{tikzpicture}
  \caption{An Undirected Brain Network.}
  \label{fig:undir_viz}
\end{minipage}
\end{figure}


We temporally partition the fMRI signal into chunks with length $L$ recorded during each cognitive process. The fMRI time series at each chunk is used to estimate a network to represent the spatio-temporal relationship among anatomic regions. Then, the cognitive process $k$ of subject $s$ is described as a consecutive list ($T_{k}^{s}$) of brain networks, formed for each chunk within time interval $[t, t+L]$, where $T_{k}^{s} = \{ G_{1}, G_{2}, \ldots, G_{C_k}\}$. Note that, $C_k$ is the number of chunks obtained for cognitive process $k$ and equals to $\lfloor N_k/L \rfloor$, where $N_k$ denotes the number of measurements recorded for cognitive process $k$. Since we obtain a different network for each duration of length $L$ for a cognitive process of length $N_k$, this approach estimates a dynamic network for the cognitive process, assuming that $N_k$ is sufficiently large. 




For a given time interval $[t, t+L]$, weights of incoming edges to vertex $v_i$ is defined by an $M $ dimensional vector, $\mathbf{\overline{a}_{i}} = [ a_{i,1}, a_{i,2} ... a_{i,M} ]$. Note that the $ith$ entry $a_{i,i} = 0$, which implies that a node does not have an edge value into itself. These edge weights define the linear dependency of activation, $b_{i,t}$, of region $r_i$ at time $t$  to the activations of the remaining regions, $b_{j,t}$ for a time interval $t' \in \{t,t+L\}$

\begin{equation}
b_{i,t'} = \sum_{j\neq i,j=1}^{M} a_{i,j} b_{j,t'} + \epsilon_{t'} = \hat{b}_{i,t'} + \epsilon_{t'}  \qquad \forall t' \in  \{t,t+L\}  
\label{eq:mesh_region_est}
\end{equation}


where $\hat{b}_{i,t'}$ is the estimated value of $b_{i,t'}$ at time $t'$ with error rate $\epsilon_{t'}$, which is the difference between the real and estimated activation. Note that  each node is connected to the rest of $M-1$  nodes each of which corresponding to anatomic regions.  





\subsection{Directed Artificial Brain Networks (dABN)}

In fully connected directed networks, we define two distinct edges between all pairs of vertices, $E = \{ e_{i,j}, e_{j,i}| v_i,v_j \in V, i \neq j \}$ where $e_{i,j}$ denotes an edge from $v_i$ to $v_j$. The weights of the edge pairs are not to be symmetrical, $a_{i,j} \neq a_{j,i}$.

The neural network we design to estimate edge weights consists of an input layer and an output layer. For every edge in the brain network, we have an equivalent weight in the neural network, such that weight between $input_{i}$ and $output_{j}$, $w_{i,j}$ is assumed to be an estimate for the weight , $a_{i,j}$ of the edge from $v_{i}$ to $v_{j}$, in the artificial brain network. 


\tikzset{%
  every neuron/.style={
    circle,
    draw,
    minimum size=0.75cm
  },
  neuron missing/.style={
    draw=none, 
    scale=2,
    text height=0.333cm,
    execute at begin node=\color{black}$\vdots$
  },
}
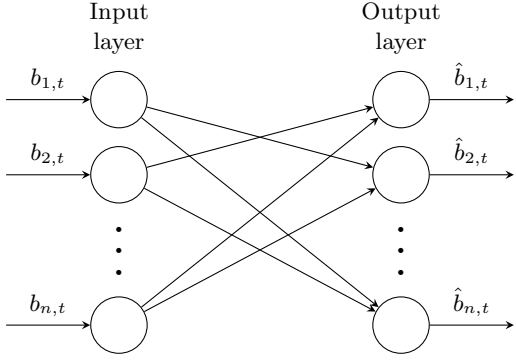
\begin{figure}
\begin{tikzpicture}[x=1.5cm, y=1cm, >=stealth]
\foreach \m/\l [count=\y] in {1,2,missing,n}
  \node [every neuron/.try, neuron \m/.try] (input-\m) at (0,2.5-\y) {};

\foreach \m [count=\y] in {1,2,missing,n}
  \node [every neuron/.try, neuron \m/.try ] (output-\m) at (2.5,2.5-\y) {};

\draw [<-] (input-1) -- ++(-1,0)
    node [above, midway] {$b_{1,t}$};
    
\draw [<-] (input-2) -- ++(-1,0)
    node [above, midway] {$b_{2,t}$};

\draw [<-] (input-n) -- ++(-1,0)
    node [above, midway] {$b_{n,t}$};

\foreach \i in {2,n}
	\draw [->] (input-1) -- (output-\i);

\foreach \i in {1,n}
	\draw [->] (input-2) -- (output-\i);

\foreach \i in {1,2}
	\draw [->] (input-n) -- (output-\i);

\draw [->] (output-1) -- ++(1,0)
  node [above, midway] {$\hat{b}_{1,t}$};

\draw [->] (output-2) -- ++(1,0)
  node [above, midway] {$\hat{b}_{2,t}$};
  
\draw [->] (output-n) -- ++(1,0)
  node [above, midway] {$\hat{b}_{n,t}$};

\foreach \l [count=\x from 0] in {Input, Output}
  \node [align=center, above] at (\x*2.5,2) {\l \\ layer};

\end{tikzpicture}
\caption{Directed Artificial Brain Network Architecture.}
\label{fig:directed_nn}
\end{figure}
We employ a regularization term $\lambda$ to increase generalization capability of the model and minimize the expected value of sum of squared error through time. Loss of an output node $output_i$ is defined as,

\begin{equation}
Loss(output_i) = E (( b_{i,t'} - \sum_{j\neq i,j=1}^{M} w_{i,j} b_{j,t'} )^2) + \lambda  \mathbf{w}^{T}_{i} \mathbf{w}_{i} 
\label{eq:dir_loss} ,
\end{equation}
where $w_{i,j}$ denotes the neural network weight between $input_i$ and $output_j$ and E(.) is the expectation operator taken over time interval [t,t+L]. For each training step of the neural network, $e$, gradient descent is applied for the optimization of the weights as in Equation \eqref{eq:dirUpdate} with empirically chosen learning rate, $\alpha$. The whole system is trained for an empirically selected number of epochs.

\begin{equation}
w_{i,j}^{(e)} \leftarrow w_{ij}^{(e-1)} - \alpha \frac{\partial Loss(output_i)}{\partial w_{i,j}} .
\label{eq:dirUpdate} 
\end{equation}

After training, the weights of neural network are assigned to edge weights of the corresponding brain graph, $ a_{i,j} \gets w_{i,j}, \forall_{i,j}$ .

\subsection{Undirected Artificial Brain Network (uABN)}

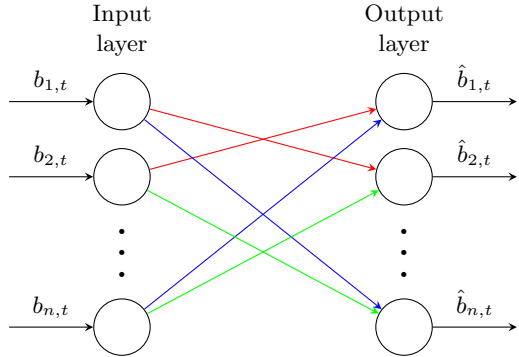
\begin{figure}
\begin{tikzpicture}[x=1.5cm, y=1cm, >=stealth]
\foreach \m/\l [count=\y] in {1,2,missing,n}
  \node [every neuron/.try, neuron \m/.try] (input-\m) at (0,2.5-\y) {};

\foreach \m [count=\y] in {1,2,missing,n}
  \node [every neuron/.try, neuron \m/.try ] (output-\m) at (2.5,2.5-\y) {};

\draw [<-] (input-1) -- ++(-1,0)
    node [above, midway] {$b_{1,t}$};
    
\draw [<-] (input-2) -- ++(-1,0)
    node [above, midway] {$b_{2,t}$};

\draw [<-] (input-n) -- ++(-1,0)
    node [above, midway] {$b_{n,t}$};

\draw [red, ->] (input-1) -- (output-2);
\draw [red, ->] (input-2) -- (output-1);

\draw [blue, ->] (input-1) -- (output-n);
\draw [blue, ->] (input-n) -- (output-1);

\draw [green, ->] (input-2) -- (output-n);
\draw [green, ->] (input-n) -- (output-2);

\draw [->] (output-1) -- ++(1,0)
  node [above, midway] {$\hat{b}_{1,t}$};

\draw [->] (output-2) -- ++(1,0)
  node [above, midway] {$\hat{b}_{2,t}$};
  
\draw [->] (output-n) -- ++(1,0)
  node [above, midway] {$\hat{b}_{n,t}$};

\foreach \l [count=\x from 0] in {Input, Output}
  \node [align=center, above] at (\x*2.5,2) {\l \\ layer};

\end{tikzpicture}
\caption{Neural Network Structure to Create Undirected Artificial Brain Networks (connections with the same colors are shared).}
\label{fig:undirected_nn}
\end{figure}

In undirected brain networks, similar to directed brain network, we define double connections between every pair of vertices $E = \{ e_{i,j}, e_{j,i}| v_i,v_j \in V, i \neq j \}$.  However, in order to make the network undirected, we must satisfy the constraint that twin (opposite) edges have the equal weights, $a_{i,j} = a_{j,i}$. In order to assure th's property in the neural network explained in the previous section, we use a weight sharing mechanism and keep the weights of the twin (opposite) edges in the neural network equal through the learning process, such that $w_{i,j} = w_{j,i}$. The proposed architecture is shown at Figure \ref{fig:undirected_nn}.

We use Equation \eqref{eq:dir_loss} for undirected Artificial Brain Networks. The weight matrix of uABN is initialized symmetrically, $w_{i,j}=w_{j,i}$ and in order to satisfy the symmetry constraint through training epochs, we define the following update rule for the weights, $w_{i,j}$ and $w_{j,i}$ at epoch $e$. 


\begin{equation}
w_{i,j}^{(e)} = w_{j,i}^{(e)} \leftarrow w_{i,j}^{(e-1)} - \frac{1}{2}  \alpha \bigg[ \frac{\partial Loss(output_i)}{\partial {w_i,j}}  + \frac{\partial Loss(output_j)}{\partial w_{i,j} } \bigg]     
\label{eq:bidirUpdate} .
\end{equation}

Again, after an empirically determined number of epochs, the weights of edges in the undirected graph is assigned to the neural network weights , $a_{i,j} \gets w_{i,j}$.

\subsection{Baseline Methods}

In this subsection, we briefly describe the popular methods that have been used to build functional connectivity graphs, in order to provide some comparison for the suggested Artificial Brain Network.

\subsubsection{Pearson Correlation}

In their work, Richiardi et al. \cite{richiardi2011decoding} defined the functional connectivity between two anatomic regions as pair-wise Pearson correlation coefficients computed between the average activations of these regions in a time interval. The edge weights are calculated by,

\begin{equation}
\rho_{b_{i,t,L}, b_{j,t,L}} = \frac{cov(\mathbf{b_{i,t,L}}, \mathbf{b_{j,t,L}})}{\sigma_{\mathbf{b_{i,t,L}} } \sigma_{\mathbf{ b_{j,t,L}} }  } ,
\end{equation}

where $\mathbf{b_{i,t,L}} = [b_{i,t}, b_{i,(t+1)}, ... , b_{i,(t+L)}]$ represents the average time series of BOLD activations of region $i$ between time $t$ and $t+L$, $cov()$ defines the covariance, and $\sigma_{s}$ represents the standard deviation of time series s. This approach assumes that the pair of similar time series represent the same cognitive process measured by fMRI signals.

\subsubsection{Closed Form Ridge Regression}


In order to generate brain networks with the method proposed in \cite{onal2015modeling}, we estimate the activation of a region from the activations of its neighboring regions in a time interval $[t, t+L]$. We minimize the loss function in Equation \ref{eq:dir_loss} using closed form solution for ridge regression. The loss function is minimized with respect to the edge weights outgoing from a vertex $v_i$, $ \overline{a}_{i} = [ a_{i,1}, a_{i,2} ... a_{i,M} ]$ and the following closed form solution of ridge regression is obtained:

\begin{equation}
\mathbf{\overline{a}_{i}} = (\mathbf{B}^{T}\mathbf{B} + \lambda \mathbf{I})^{-1}\mathbf{B}^{T} \mathbf{b_{i,t,L}} ,
\end{equation}
where $\mathbf{B}$ is an $L \times (M-1)$ matrix, whose columns consist of the average BOLD activations of anatomic regions except for the region $r_i$ in the time interval $[t, t+L]$ such that column $j$ of matrix $\mathbf{B}$ is $\mathbf{b_{j,t,L}}$ . $\lambda \in R$ represents the regularization parameter. 

\section{Experiments \& Results}

In order to examine the representation power of the suggested Artificial Brain Networks,  we compare them with the baseline methods, presented in the previous subsection, on two different fMRI dataset. The comparison is done by measuring the cognitive task classification performances  of all the models.

\subsection{Human Connectome Project (HCP) Experiment}

In Human Connectome Project dataset, 808 subjects attended 7 sessions of fMRI scanning in each of which the subjects were required to complete a different cognitive task with various durations, namely, Emotion Processing, Gambling, Language, Motor, Relational Processing, Social Cognition, and Working Memory. We aim to discriminate these 7 tasks using the edge weights of the formed brain graphs. 


In the experiments, the learning rate $\alpha$ was empirically chosen as $\alpha = 10^{-5}$ for both dABN and uABN and window size is chosen as $L = 40$. We tested the directed and undirected Artificial Brain Networks and Ridge Regression method using various $\lambda$ values. Since computation of Pearson correlation does not require any hyper parameter  estimation, a single result is obtained for the Pearson correlation method. 


After estimating the Artificial Brain Networks and forming the feature vectors from edge weights of the brain networks, we performed within-subject and across-subject experiments using Support Vector Machines with linear kernel. During the within-subjects experiments, we performed 3-fold cross validation using only the samples of a single subject. Table \ref{table:within_itir} shows the average of within-subject experiment results over 807 subjects, when the classification is performed using a single subject brain network of 7 tasks. During the across-subject experiments, we performed 3-fold cross validation using the samples obtained from 807 subjects. For each fold we employed the samples from 538 subjects to train and 269 subject to test the classifier. Table \ref{table:across_itir} shows the across-subject experiment results. 

\begin{table}[]
\centering
\caption{Within-Subject Performances of Brain Networks on HCP Dataset.}
\label{table:within_itir}
\begin{tabular}{l
@{\hspace*{4mm}} c
@{\hspace*{\lengtha}}c
@{\hspace*{\lengthb}}c
@{\hspace*{4mm}}c
@{\hspace*{\lengtha}}c
@{\hspace*{\lengthb}}c
@{\hspace*{4mm}}c
@{\hspace*{\lengtha}}c
@{\hspace*{\lengthb}}c
@{\hspace*{4mm}}c
@{\hspace*{\lengtha}}c
@{\hspace*{\lengthb}}c}
\toprule
 &  & \multicolumn{2}{c}{Pearson Corr.} &  & \multicolumn{2}{c}{Ridge Reg.} &  & \multicolumn{2}{c}{dABN} &  & \multicolumn{2}{c}{uABN} \\
\cmidrule{3-4} \cmidrule{6-7} \cmidrule{9-10} \cmidrule{12-13}
$\lambda$ &  & Mean & Std &  & Mean & Std &  & Mean & Std &  & Mean & Std \\
\midrule
0 &  & 0.7194 & 0.16 &  & - & - &  & \textbf{0.7435} & 0.13 &  & 0.5918 & 0.13 \\
32 &  & 0.7194 & 0.16 &  & 0.7957 & 0.11 &  & \textbf{0.9133} & 0.08 &  & \textbf{0.913} & 0.08 \\
64 &  & 0.7194 & 0.16 &  & 0.8304 & 0.11 &  & \textbf{0.9406} & 0.07 &  & \textbf{0.9402} & 0.07 \\
128 &  & 0.7194 & 0.16 &  & 0.8377 & 0.11 &  & \textbf{0.9463} & 0.06 &  & \textbf{0.9462} & 0.07 \\
256 &  & 0.7194 & 0.16 &  & 0.8119 & 0.12 &  & \textbf{0.9313} & 0.08 &  & \textbf{0.9307} & 0.08 \\
512 &  & 0.7194 & 0.16 &  & 0.7462 & 0.13 &  & \textbf{0.8852} & 0.1 &  & \textbf{0.8849} & 0.1 \\
\bottomrule
\end{tabular}
\end{table}

\begin{table}[]
\centering
\caption{Across-Subject Performances of Brain Networks on HCP Dataset.}
\label{table:across_itir}
\begin{tabular}{l
@{\hspace*{4mm}} c
@{\hspace*{\lengtha}}c
@{\hspace*{\lengthb}}c
@{\hspace*{4mm}}c
@{\hspace*{\lengtha}}c
@{\hspace*{\lengthb}}c
@{\hspace*{4mm}}c
@{\hspace*{\lengtha}}c
@{\hspace*{\lengthb}}c
@{\hspace*{4mm}}c
@{\hspace*{\lengtha}}c
@{\hspace*{\lengthb}}c}
\toprule
 &  & \multicolumn{2}{c}{Pearson Corr.} &  & \multicolumn{2}{c}{Ridge Reg.} &  & \multicolumn{2}{c}{dABN} &  & \multicolumn{2}{c}{uABN} \\
\cmidrule{3-4} \cmidrule{6-7} \cmidrule{9-10} \cmidrule{12-13}
$\lambda$ &  & Mean & Std &  & Mean & Std &  & Mean & Std &  & Mean & Std \\
\midrule
0  &  & \textbf{0.7524} & 0.01 &  &   -    &   -  &  & 0.6654 & 0.01 &  & 0.5681 & 0.01 \\
32 &  & 0.7524 & 0.01 &  & 0.8027 & 0.01 &  & \textbf{0.8153} & 0.00 &  & 0.8123 & 0.00 \\
64 &  & 0.7524 & 0.01 &  & 0.8223 & 0.00 &  & \textbf{0.8312} & 0.01 &  & 0.8297 & 0.01\\
128&  & 0.7524 & 0.01 &  & 0.8370 & 0.01 &  & \textbf{0.8401} & 0.01 &  & 0.8393 & 0.01 \\
256&  & 0.7524 & 0.01 &  & \textbf{0.8461} & 0.01 &  & 0.8410 & 0.01 &  & 0.8406 & 0.00 \\
512&  & 0.7524 & 0.01 &  & \textbf{0.8466} & 0.01 &  & 0.8357 & 0.01 &  & 0.8357 & 0.01 \\
\bottomrule
\end{tabular}
\end{table}

Table \ref{table:within_itir} shows that in within subject experiments our methods, dABN and uABN, have the best performances in classifying the cognitive task under different $\lambda$ values, furthermore performances of directed networks are slightly better than undirected ones. It can be observed that as $\lambda$ increases, generalization of our models also increase up to $\lambda=128$. 

Table \ref{table:across_itir} shows that our methods outperforms the others within a range of lambdas, $\lambda = \{ 32, 64, 128 \}$. Pearson Correlation results in the best accuracy when no regularization is applied to Artificial Brain Networks. Closed Form Ridge Regression solution offers more discriminative power in higher $\lambda$ values. 

\subsection{Tower of London(TOL) Experiment}

We also test the validity of the suggested Artificial Brain Network  on a relatively more difficult fMRI dataset, recorded when the subjects solve Tower of London (TOL) problem. TOL is a puzzle game which has been used to study complex problem solving tasks in human brain. TOL dataset used in our experiments contains fMRI measurements of 18 subjects attending 4 session of problem solving experiment. In the fMRI experiments, subjects were asked to solve 18 different puzzles on computerized version of TOL problem \cite{newman2009fmri}. There are two labeled subtask of problem solving with varying time periods namely, planning and execution phases. 

As the nature of the data is not compatible with a sliding window approach and the dimensionality is too high for a computational model, in the study of Alchihabi et al. \cite{alchihabi2018decoding}, a series of preprocessing steps were suggested for the TOL dataset. In this study, we employ the first two steps of their pipeline. In the first step called \textit{voxel selection and regrouping}, a feature selection method is applied on time series of voxels to select the "important" ones. Then, the activations of the selected voxels in the same region are averaged to obtain the activity of corresponding region. As a result, a more informative and lower dimensional representation is achieved. In the second step, bi-cubic spline interpolation is applied to every consecutive brain volumes and a number of new brain volumes are inserted between two brain volumes  to increase temporal resolution. For the details of interpolation, refer to \cite{alchihabi2018decoding}. In this study, the optimal number of volumes inserted between two consecutive brain volumes are found empirically and it is set to 4. Therefore, the time resolution of the data is increased four times.

We applied the above-mentioned preprocessing steps to all of the 72 sessions in the dataset. After the voxel selection phase, number of regions containing selected voxels is much less than 116 regions. Note that, we discard regions located in Cerebellum and Vermis. Window size for this dataset was set to $L = 5$, since there are  at least 5 measurements for every sub-phase after the interpolation. The neural network parameters used in our experiments are $\alpha=10^{-6}$ and $\# epochs = 10$. Table \ref{table:within_tol} shows the mean and standard deviation of classification accuracies obtained with our method and the base-line methods. Similar to HCP experiments, we slided non-overlapping windows on the measurements and we performed 3-fold cross validation during TOL experiments.

\begin{table}[]
\centering
\caption{Across-Subject Performances of Mesh Networks on TOL Dataset.}
\label{table:within_tol}
\begin{tabular}{l
@{\hspace*{4mm}} c
@{\hspace*{\lengtha}}c
@{\hspace*{\lengthb}}c
@{\hspace*{4mm}}c
@{\hspace*{\lengtha}}c
@{\hspace*{\lengthb}}c
@{\hspace*{4mm}}c
@{\hspace*{\lengtha}}c
@{\hspace*{\lengthb}}c
@{\hspace*{4mm}}c
@{\hspace*{\lengtha}}c
@{\hspace*{\lengthb}}c}
\toprule
 &  & \multicolumn{2}{c}{Pearson Corr.} &  & \multicolumn{2}{c}{Ridge Reg.} &  & \multicolumn{2}{c}{dABN} &  & \multicolumn{2}{c}{uABN} \\
\cmidrule{3-4} \cmidrule{6-7} \cmidrule{9-10} \cmidrule{12-13}
$\lambda$ &  & Mean & Std &  & Mean & Std &  & Mean & Std &  & Mean & Std \\
\midrule
0  &  & 0.6119 & 0.09 &  &   -    &   -  &  & \textbf{0.8914} & 0.11 &  & 0.8499 & 0.12 \\
32 &  & 0.6119 & 0.09 &  & 0.6688 & 0.10 &  & \textbf{0.8913} & 0.11 &  & 0.8499 & 0.12 \\
64 &  & 0.6119 & 0.09 &  & 0.6651 & 0.10 &  & \textbf{0.8914} & 0.11 &  & 0.8499 & 0.12 \\
128&  & 0.6119 & 0.09 &  & 0.6679 & 0.10 &  & \textbf{0.8906} & 0.11 &  & 0.8499 & 0.12 \\
256&  & 0.6119 & 0.09 &  & 0.6685 & 0.10 &  & \textbf{0.8905} & 0.11 &  & 0.8500 & 0.12 \\
512&  & 0.6119 & 0.09 &  & 0.6705 & 0.10 &  & \textbf{0.8912} & 0.11 &  & 0.8498 & 0.12 \\
\bottomrule
\end{tabular}
\end{table}

Table \ref{table:within_tol} shows that using Artificial Brain Networks gives better performances than using Pearson Correlation and Closed Form Ridge Regression methods in classifying sub-phases of complex problem solving under various regularization parameters. We observe that decoding performances of directed brain networks outperforms those of undirected brain networks.


\section{Discussion and Future Work}

In this study, we introduce a network representation of fMRI signals, recorded when the subjects perform a cognitive task. We show that the suggested Artificial Brain Network estimated from the average activations of anatomic regions using an artificial neural network leads to a powerful representation to discriminate cognitive processes. Compared to the brain networks obtained by ridge regression,   the suggested Artificial Brain Network achieves  more discriminative features. The success of the suggested brain network can be attributed to the iterative nature of the neural network algorithms to optimize the loss function, which avoids the singularity problems of Ridge Regression.

In most of the studies, it is customary  to represent functional brain connectivities as an undirected graphs. However, in this study, we observe that the directed network representations capture more discriminative features compared to the undirected ones in brain decoding problems. 

In this study, we  consider complete brain graphs where all regions are assumed to have connections to each other. A sparser brain representation can be computationally more efficient and neuro-scientifically more accurate. As a future work, we aim to estimate more efficient brain network representations by employing some sparsity parameters in the artificial neural networks.


It is well-known that brain processes the information in various frequency bands. \cite{richiardi2011decoding} and \cite{ertugrul2016hierarchical} applied discrete wavelet transform before creating connectivity graphs. A similar approach can be taken for a more complete temporal information in brain decoding problems.

\section{Acknowledgment}
The work is supported by TUBITAK (Scientific and Technological Research Council of Turkey) under the grant No: 116E091. We also thank Sharlene Newman, from Indiana University,  for providing us the TOL dataset.

\bibliographystyle{splncs04}
\bibliography{strings}

\end{document}